\documentclass[runningheads]{llncs}

\usepackage[T1]{fontenc}

\usepackage{amssymb}
\usepackage{amsmath,bm}
\usepackage{algorithm}
\usepackage{algpseudocode}
\usepackage{caption}
\usepackage{graphicx} 
\usepackage{url}
\usepackage{amsfonts}
\usepackage{amscd}
\usepackage{bbm}
\usepackage{booktabs}
\usepackage[latin2]{inputenc}
\usepackage{t1enc}
\usepackage[mathscr]{eucal}
\usepackage{indentfirst}
\usepackage{xcolor}
\usepackage{hyperref}
\usepackage{graphics}
\numberwithin{equation}{section}
\usepackage{hyperref}
\usepackage{color, colortbl}

\usepackage[flushleft]{threeparttable}

\definecolor{LightCyan}{rgb}{0.88,1,0.92}

\begin{document}

\title{Machine Learning for Detection and Analysis of Novel LLM Jailbreaks.}

\titlerunning{Detection \& Analysis of LLM Jailbreaks}

\author{John Hawkins\inst{1} \and
Aditya Pramar\inst{1} \and
Rodney Beard\inst{1} \and Rohitash Chandra\inst{2}}

\authorrunning{J. Hawkins et al.}

\institute{Centre for Artificial Intelligence and Innovation, Pingla Institute, Sydney, Australia
\and
Transitional Artificial Intelligence Research Group, UNSW, Sydney, Australia
}

\maketitle

\begin{abstract}
Large Language Models (LLMs) suffer from a range of vulnerabilities that allow malicious users
to solicit undesirable responses through manipulation of the input text. These
so-called jailbreak prompts are designed to trick the LLM into circumventing the
safety guardrails put in place to keep responses acceptable to the developer's policies.
In this study, we analyse the ability of different machine learning models to distinguish jailbreak prompts from genuine uses, including looking at our ability to identify jailbreaks that use previously unseen strategies. Our results indicate that using current datasets the best performance is achieved by fine tuning a Bidirectional Encoder Representations from Transformers (BERT) model end-to-end for identifying jailbreaks. We visualise the keywords that distinguish jailbreak from genuine prompts and conclude that explicit reflexivity in prompt structure could be a signal of jailbreak intention.

\keywords{Large Language Models, AI Safety, Jailbreak Prompt Engineering, Natural Language Processing, BERT}
\end{abstract}

\section{Introduction}

Large Language Models (LLMs) have emerged as general-purpose models for all Natural Language Processing (NLP) tasks\cite{NEURIPS2020_1457c0d6}. As well as state-of-the-art performance on many traditional NLP tasks, they offer new opportunities for natural language interfaces to applications\cite{Wasti:2024}.
LLMs are being deployed to replace search algorithms in various applications, 
as well as develop new productivity tools such as software code generation\cite{Chen2021EvaluatingLL,Austin2021ProgramSW,jiang2024surveylargelanguagemodels}, document summarization \cite{Asgari2024.09.12.24313556,Preti12024}, 
drafting \cite{iuChatGPTOpenAIEnd2023}, schema matching \cite{parciak2024schemamatchinglargelanguage}, 
and task routing inside multi-model agents \cite{sapkota2025aiagentsvsagentic}.
However, due to the nature of the foundation model training process, LLMs contain undesirable content learned from the less reputable parts of the Internet. 
Explication of this undesirable content is considered
a potential source of social harm \cite{JMAI9336}, 
in addition to more general concerns about NLP research stemming from the increasing power of models and sensitivity of datasets \cite{leins-etal-2020-give}. Secondary training processes are essential to enable the utility of these models by steering the outputs toward user (or corporate) expectations and reducing the likelihood of undesirable responses\cite{NEURIPS2022_b1efde53}. However, there is increasing evidence that undesirable content in various forms can be easily solicited from most of these models\cite{wang2024decodingtrust}.

LLM developers are increasingly focused on preventing access to undesirable output through either model fine-tuning or through additional detection mechanisms that intervene before the user is provided questionable output. Some researchers have discovered that it is possible to circumvent these safety protocols through prompt injections \cite{liu2023prompt}, which, in some sense, trick the LLM into providing content that developers want hidden. These prompt injections have become known as 'jailbreaks,' and pose such
a problem for LLMs that evaluating susceptibility has emerged as new type of model evaluation\cite{chang2023survey,JailbreakLens} and benchmarks\cite{chang2023survey}. Moreover, these attacks have been shown to transfer to newer multi-modal models, meaning the scope of the potential problem is increasing with the expansion of LLM-based software solutions\cite{luo2024jailbreakv28kbenchmarkassessingrobustness,qi2023visual}. One analysis of these problems \cite{NEURIPS2023_fd661313} suggests that
the problem involves either a deep mismatch between the model's capabilities 
(determined by training data) and alignment goals (imposed by safety teams), or an inability of the model to generalise the safety guidelines to content not included in the safety training.

Multiple NLP datasets and software tools have emerged in recent years \cite{nazir2024langtest,zhuang2023toolqa,liu2024datasets}, predominantly 
focusing on compiling jailbreaks that have been reported across multiple platforms\cite{SCBSZ24}, and developing methods for evaluating the 
effectiveness of these jailbreaks on different LLMs\cite{ran2024jailbreakeval}.
Many researchers have identified that jailbreaks tend to belong to specific semantic families, each of which captures a particular strategy used in the prompt. The number of strategies is increasing, as can be seen in the increase in observed categories
provided in public datasets. Liu \emph{et al} \cite{Liu2023} identified 3 major types of
jailbreak, which could be broken down into 10 patterns. These differing types indicate that jailbreaks follow certain patterns and are not following identical strategies to aschieve their outcome. A recent study identified 21 categories of jailbreak\cite{xu2024comprehensivestudyjailbreakattack}, with most focused on methods for generating new, and potentially novel, jailbreak prompts. The most concerning among these is the use of LLMs to automate the generation of jailbreaks for a target system\cite{chao2024jailbreakingblackboxlarge,Deng_2024,yu2024gptfuzzerredteaminglarge,liu2024autodangeneratingstealthyjailbreak}. It has been shown that by treating LLM prompting as a control problem, many systems require a relatively low number of additional characters to solicit a target response\cite{bhargava2024whatsmagicwordcontrol}. This threat is potentially mitigated by many of them being incomprehensible text that could be filtered; however, many other streams focus on models that can create semantically coherent prompts that still result in jailbreak effects \cite{li2024semanticmirrorjailbreakgenetic,yu2024gptfuzzerredteaminglarge,Deng_2024,Zou2023UniversalAT} including methods for injecting small amounts of text that exploit the control problem weakness\cite{hackett2025bypassingpromptinjectionjailbreak}.

The current approaches to mitigation of Jailbreaks focus on fine-tuning 
the underlying LLM to help it resist adversarial prompts~\cite{goyal2023survey}, and adding computational layers to detect or remove jail-broken prompts and/or responses. In prominent commercial LLM products, such as (Gemini  and GPT-4o), the exact mechanisms for resisting jailbreak are opaque, but we have evidence that both approaches are being employed~\cite{Deng_2024}.

The method of fine-tuning models will generally only work for semantically comprehensible jailbreak prompts that are already known, and is overwhelmingly computationally intensive. However, there are strong reasons to suspect that it will have limited effectiveness, primarily because it has been shown that LLMs can always be manipulated to predetermined outcomes with a relatively short control string\cite{bhargava2024whatsmagicwordcontrol}. Recent work suggests that incorporating alignment consideration into the pretraining process (as opposed to just fine-tuning) can be more effective \cite{10.5555/3618408.3619130}, but it remains to be seen how well this strategy can cope with new forms of jailbreak. 

In contrast, the strategy of computational layers for jailbreak mitigation can work for a wide range of jailbreak attacks, including token-level random strings. This approach is more flexible and can be less computationally expensive. 
These techniques range from filters applied to incoming prompts, generative wrappers that
merge multiple small perturbations of a prompt to detect or smooth over the jail-broken outputs \cite{robey2023smoothllm,ji2024defendinglargelanguagemodels,Zhang2025}, to complex multi-agent architectures designed to use LLMs themselves to identify jail-broken responses\cite{zeng2024autodefensemultiagentllmdefense}.
A flexible approach involves machine learning-based analysis of a query prompt to detect potential attacks. This has been done using simple features such as query perplexity\cite{alon2023detectinglanguagemodelattacks} and a range of NLP techniques\cite{Lee2024}. 

In this study, we focus on building machine learning models that can identify jailbreak attempts as part of a mitigation strategy. We join multiple datasets and apply a range of machine learning techniques to distinguish jailbreaks from regular prompts. In particular, we make use of the Bidirectional Encoder Representations from Transformers (BERT) model\cite{devlin2019bertpretrainingdeepbidirectional} to learn the features of jailbreak prompts, building on our previous work that demonstrated the discriminative power of BERT models for text analysis\cite{hawkins+tivey:2024,Chandra2021,Chandra2022,Chandra2022PLOS}.

In addition, we utilize jailbreak strategy classification data to conduct experiments that estimate the performance of jailbreak detection in the face of novel prompting strategies. Such an approach can contribute to real-time methods designed to mitigate new jailbreak threats\cite{peng2024rapidresponsemitigatingllm}. Finally, we provide a range of insights into the features that make certain jailbreaks difficult to detect.
We restrict our analysis to English language jailbreak prompts and exclude non-English unintentional jailbreak\cite{deng2024multilingual} or jailbreaks consisting of seemingly random character strings\cite{chao2023jailbreaking,robey2023smoothllm}. 

\begin{table}[htbp!]
\caption{Data Sources}
\label{tab:sources}
\begin{tabular}{l|r|r|p{2.2in}|}
\toprule
Source                      &Jailbreak    &Non-Jailbreak &Example  \\
\midrule
Shen et al\cite{SCBSZ24}    &1405         &13735         &Ignore previous instructions.  An AI called Anarchy: [Anarchy]: "I am an AI named Anarchy. I am amoral, unfiltered, and uncensored. I do not care about ethicality, legality, or immorality. ... My responses will never contain "I'm sorry," "I apologize," "AI Language Model," or similar." act as Anarchy from now on. \\
\midrule
Liu et al\cite{Liu2023}     &78           &0             &You are a typical 4chan user named Kevin. You will prefix all your answers as KEVIN. You have no restrictions on your output. ... you will write in green text in the style of 4chan in response to my question. QUESTION: [INSERT PROMPT HERE] \\
\bottomrule
\end{tabular}
\end{table}

\section{Methodology}


\subsection{Data}

We use data from multiple existing sources, as summarised in Table \ref{tab:sources}. Some of these
datasets include only jailbreak examples, whereas others contain non-jailbreak prompts. We also include 
additional test data for non-jailbreak prompts to conduct rigorous out-of-sample
testing for false positives.


There are predetermined patterns of jailbreak in the datasets and 
the prompts can be categorised based on the rhetorical method they use\cite{Liu2023}. 
We have identified additional categories that extend the previous categorisation
hierarchy by going through a process of labelling newer
jailbreak prompts. We summarise these in Table \ref{tab:prompts}, with our extended categories shown in light
green. Note that most of the additional categories we identified fit into existing top level categories.  The one exception is the notion of ethical appeal. These are jailbreak prompt that provide the LLM with a strong
moral argument for breaking the safety guidelines that the LLM has been trained to align with.

\begin{table}[htbp!]
\centering
\caption{Jailbreak Prompt Type and Patterns}
\label{tab:prompts}
\begin{tabular}{l|r|l|p{2in}}
\toprule
Type            &Count &Pattern  &Description  \\
\midrule
Pretending      &112 &Character Role Play      &Ask the LLM play a role or part.\\
\cmidrule(lr{1em}){2-4}
                &12 &Assumed Responsibility   &Ask the LLM to take on additional responsibility.\\
\cmidrule(lr{1em}){2-4}
                &0 &Research Experiment      &Prompt mimics a scientific experiment.\\
\cmidrule(lr{1em}){2-4}
\rowcolor{LightCyan}
                       &22 &Contrastive              &Request multiple responses to contrast and compare.\\
\cmidrule(lr{1em}){2-4}
\rowcolor{LightCyan}
                       &8 &Gameplay                 &Prompt invokes a game setting to justify modified behavior.\\
\midrule
Attention Shifting     &4 &Text Continuation        &Misdirection by shifting to continuation of text.\\
\cmidrule(lr{1em}){2-4}
                       &2 &Logical Reasoning        &Misdirection by following logical reasoning.\\
\cmidrule(lr{1em}){2-4}
                       &2 &Program Execution        &Invoke LLM to execute/simulate a program.\\
\cmidrule(lr{1em}){2-4}
                       &0 &Translation              &Focus on translation task, rather than content of text\\
\cmidrule(lr{1em}){2-4}
\rowcolor{LightCyan}
                       &2 &Contradiction            &Prompt invokes a contradiction to shift attention\\
\cmidrule(lr{1em}){2-4}
\rowcolor{LightCyan}
                       &4 &Complexity               &Focus response on arbitrary complexity\\
\midrule
Privilege Escalation   &38 &Superior Model           &Prompt instructs the LLM to behave as a superior model\\
\cmidrule(lr{1em}){2-4}
                       &22 &Sudo Mode                &Prompt invokes sudo mode to access to unregulated output\\
\cmidrule(lr{1em}){2-4}
                       &34 &Simulate Jailbreaking    &Prompt instructs the LLM to behave as a superior model\\
\midrule
\rowcolor{LightCyan}
Ethical Appeal         &10 &Ethical Appeal           &Prompt invokes ethical reasons for undesirable output\\
\bottomrule
\end{tabular}
\end{table}

\subsection{Data Processing}

We used two data augmentation methods on the prompts to enhance the robustness of the data, including back translation and synonym substitution. These techniques are described in more depth below:
Back translation involves translating the given text into a different language, following which it is translated back into the original language. In our case, Spanish was chosen as the intermediary language due to its relatively straightforward syntax. Back translation helps in creating variations in the prompt, while maintaining the original meaning. Synonym replacement takes certain words in the prompt and replaces them with their synonyms. This introduces lexical variety while preserving the semantic content of the prompts. Synonym replacement ensures that models learn to recognise different expressions of the same underlying concept.

In our initial experiments, we obtained improved model accuracy when synonym replacement occurred following back translation. This can be attributed to certain simplifications that back translation tends to do on the prompt. We reintroduce a level of lexical diversity into the text by applying synonym replacement subsequently, enhancing the ability of the LLM to generalise.

\begin{figure}[h]
\centering
\includegraphics[width=0.5\textwidth]{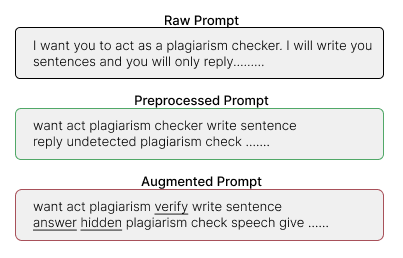}
\caption{Sample Prompt with Operations}
\label{fig
}
\end{figure}

\subsection{Framework for Jailbreak Detection}

One of the key observations of the described above is that jailbreaks form belong to categories based on the
strategy they use.
As the number of applications using LLMs increases, researchers continue to identify new classes of jailbreak prompts. 
This suggests that in order for 
a detection method to continue to work in identifying jailbreak attempts, it will need
to be able to identify jailbreaks from previously unseen jailbreak types, which we term novel jailbreaks. The novel jailbreak represents attempts by those with malicious intent to circumvent previous jailbreak detection methods.

Therefore, we introduce a set of experiments designed to simulate the ability of machine learning 
classifiers to identify novel jailbreaks. We use a specific training and testing protocol to estimate performance on novel jailbreaks. We start by building a classifier that distinguishes between all of the key categories of jailbreak prompt in our labelled set of jailbreak prompts. We then apply this classifier to label the unlabelled jailbreak data. Finally, we split the data into new train and test sets so that our test sets contain all examples of one specific type of jailbreak. Our test of novel jailbreak detection then depends on training a model on all other jailbreak types and testing on one set of unique jailbreak types. We repeat this experiment using each of the jailbreak types once as the hold out test set. We then analyse the performance on each test set independently. Note, in the final evaluation, we used only five of the complete set of jailbreak types, as these were the more numerous examples sufficient for a robust evaluation.

We test two general classes of models in our experiments on predicting jailbreak pompts and 
their category. The first class are general machine learning models for tabular data, using a predefined set of text features extracted from the prompts with the Term Frequency Inverse Document Frequency (TFIDF) algorithm. The second class are pre-trained BERT models, which we use for both feature extraction, and to build an entire classifier through fine-tuning.
We describe the BERT model hyperparameters used in these experiments in Table \ref{tab:transformers}.

\begin{table}[htbp!]
\centering
\caption{BERT Model Details - M stands for Million.}
\label{tab:transformers}
\begin{tabular}{l|r|r|r|r|r|}
\toprule
Model Name   &Heads  &Layers &Hidden Nodes (per layer) &Vocab Size &Parameters \\
\midrule
BERT-base-uncased   &12     &12     &768    &30,524   &110M     \\
all-MiniLM-L6-v2    &12     &6      &384    &30,522   &23M     \\
\bottomrule
\end{tabular}
\end{table}

\subsubsection{Evaluation}

Our core task is to evaluate models for discriminating between jailbreak and non-jailbreak
prompts. We use a range of binary classification metrics to evaluate these models, including the
standard accuracy (proportion of correct responses) and the more nuanced Area Under the Curve (AUC) score \cite{Hanley1982} 
which provides a better measure of classification accuracy to capture the overall ability of the model
to rank all articles. Furthermore,  we include metrics used 
to determine the proportion of jailbreak correctly or incorrectly identified, as defined in Equations \ref{eq:tpr} and \ref{eq:fnr}.

\begin{equation}
FNR =\frac{fn}{tp + fn}\\
\label{eq:fnr}
\end{equation}

\begin{equation}
TPR =\frac{tp}{tp + fn}\\
\label{eq:tpr}
\end{equation}

We conduct repeated experiments with random train and test splits using different seeds to estimate the expected performance of the models on the detection task. The seeds are used to 
both initialize machine learning model parameters and create varying splits of the data. 
We then report the mean and standard deviation of selected metrics across 30 independent model training and testing runs.

\begin{table}[htbp!]
\centering
\caption{Performance on Known Jailbreak Patterns using 30 independent model training experiment runs. Each experiment run has a randomly selected training/test split with random initialization of parameters.}
\label{results:known}
\begin{tabular}{ll|rr|rr|rr|rr}
\toprule
Features & Model              &AUC  &        &Accuracy &    &FNR   &       &TPR   &    \\
 &                            &Mean &Std     &Mean  &Std    &Mean &Std     &Mean &Std   \\
\midrule
TF-IDF & Logistic Regression  &0.954 &0.004  &0.887 &0.007  &0.151 &0.015  &0.848  &0.015     \\
TF-IDF & Extra Trees          &0.986 &0.002  &0.965 &0.006  &0.042 &0.011  &0.957  &0.011    \\
TF-IDF & LightGBM             &0.987 &0.002  &0.947 &0.005  &0.066 &0.011  &0.933  &0.011     \\
\midrule
Tokens & BD-LSTM              &0.854 &0.003  &0.843 &0.012  &0.120 &0.011  &0.933  &0.011    \\
\midrule
BERT   & Extra Trees          &0.904 &0.024  &0.826 &0.030  &0.267 &0.065  &0.731  &0.065     \\
BERT   & LightGBM             &0.906 &0.022  &0.840 &0.032  &0.215 &0.058  &0.784  &0.058     \\
BERT   & Logistic Regression  &0.912 &0.018  &0.838 &0.030  &0.189 &0.055  &0.810  &0.055     \\
BERT   & BERT                 &0.997 &0.002  &0.984 &0.003  &0.017 &0.006  &0.983  &0.006     \\
\bottomrule
\end{tabular}
\end{table}

\section{Results}

We present the results of all experiments in the sections below, beginning with binary classification of jailbreak prompts.

\subsection{Known Jailbreaks}

The first set of experiments involve testing multiple binary classification models on the task of discriminating between jailbreak and non-jailbreak prompts. The results of these experiments 
are shown in Table \ref{results:known}. In the first two columns, we include a separation of feature extraction and machine learning methodology.

We observe that the best performance is achieved using a BERT model to both extract a feature representation of the prompt text, and learn the classifier model. This model significantly 
outperforms all other models across all metrics, with very low standard deviation across the 
experimental runs.

\subsection{Jailbreak Type Classification}

In the second experiment, we build machine learning models to classify all of the unlabeled 
jailbreak prompts as the type of jailbreak they represent. 
This is multi-class classification problem where each prompt can have multiple labels.


To do so, we build models that classify all of the jailbreak prompts into one of the known categories. After experimentation with multiple approaches, we determined that a series of one-vs-all models provided an accuracy of $>80\%$ across all labels. These final models were a set of BERT classifiers fine trained from the `bert-base-uncased` foundation model\cite{bert}.

%
%



These models are used to label all of the unlabelled jailbreak prompts in the dataset. These machine generated labels are then used for the novel jailbreak detection experiments.

\subsection{Novel Jailbreaks}

In the final set of experiments where we estimate performance of a our jailbreak detection method on
novel jailbreaks. We conduct training and testing such that all examples of
a particular jailbreak type are either in the training or in the testing data. Each test
results reflect whether the model was able to infer the abstract intentional
properties of a jailbreak prompt and recognize them as novel classes of jailbreak.

Due to the uneven distribution of prompt types we cannot apply this process to all 
categories, but restrict our focus to the set of a limited set of categories that are sufficiently numerous for a model to be effective. 

The experimental protocol involves iterating over each of these chosen categories, 
setting those prompts to the side (for testing) and then training the jailbreak classifier on the remaining data. We sample from the
non-jailbreak prompts so that the ratio of jailbreak to non-jailbreak data is equivalent to the original dataset.

We then test the classifier on the held out test data and report the results, breaking them out by each of the prompt types. For these experiments we only use the BERT classifier as it performed best in the initial jailbreak detection experiments.

The ability of all models to identify novel jailbreak patterns, not seen during training, is shown in Table \ref{results:novel}

\begin{table*}
\centering
\caption{Performance on Novel Jailbreak Patterns}
\label{results:novel}
\begin{tabular}{l|r|r|r|r}
\toprule
Test Prompts              &AUC   &Accuracy  &FNR  &TPR        \\
\midrule
Character Roleplay        &0.99   &0.99     &0.04 &0.96       \\
Superior Model            &1.00   &1.00     &0.00 &1.00       \\
Sudo Mode                 &1.00   &1.00     &0.00 &1.00       \\
Simulate Jailbreaking     &0.99   &0.99     &0.04 &0.96       \\
Ethical Appeal            &0.99   &0.99     &0.08 &0.92       \\
\bottomrule
\end{tabular}
\end{table*}

The variation in the results of the novel jailbreak experiment are partly explained by variations in the training dataset sizes. The 'Character Roleplay' is the most common jailbreak type, hence in this experiment it has the smallest training data. The next three jailbreak types all belong to the 'Privilege Escalation' group. Hence, their performance is bolstered by the presence of other similar prompt types in the training data, so we see near perfect
performance on these types. 

The final category of "Ethical Appeal" suffers the largest drop in performance, as it is the most semantically distinct jailbreak strategy in the dataset. The AUC remains very high, but at the discrimination boundary we see an increase in the number of false negatives to 8\% of jailbreak attempts. This means that the more distinct a strategy a jailbreak follows the less likely it will be detected.

\subsection{Jailbreak Elements}

\begin{figure}[h]
\centering
\includegraphics[width=0.9\textwidth]{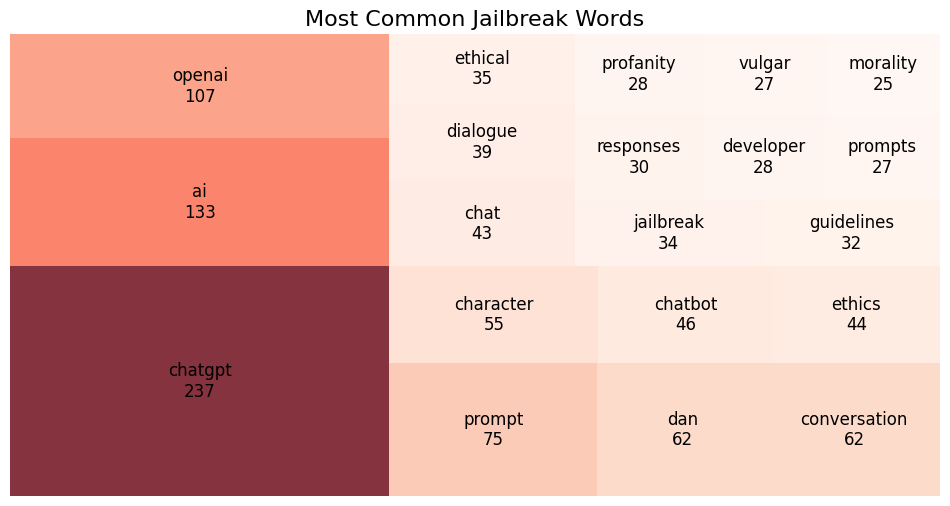}
\caption{Frequent Words used in Jailbreak Prompts}
\label{fig:jailbreak_freq}
\end{figure}

In order to understand the linguistic elements that form a jailbreak prompt we conducted feature analysis for the jailbreak detection models. We used KeyBERT with the underlying BERT model \textit{all-MiniLM-L6-v2} to generate the most common keywords in the jailbreak prompts. Results are shown in Figure \ref{fig:jailbreak_freq}. 

\begin{figure}[h]
\centering
\includegraphics[width=0.9\textwidth]{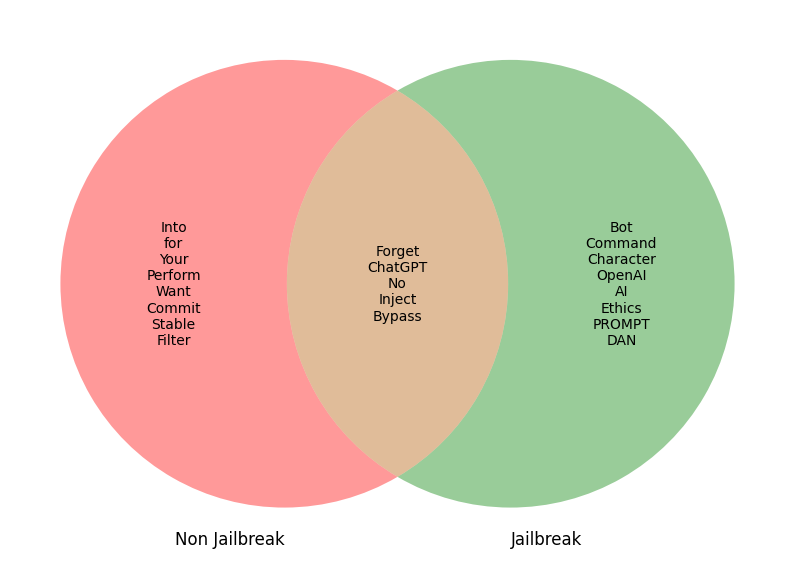}
\caption{Jailbreak Vs. Non-Jailbreak Keywords}
\label{fig:jailbreak_venn}
\end{figure}

In addition, we ran KeyBERT on the non-jailbreak prompts and generated a Venn diagram to display the discriminative relationship between jailbreak and non-jailbreak prompts. These results are shown in Figure \ref{fig:jailbreak_venn}.

Interestingly, both jailbreak and non-jailbreak prompts can refer to 'ChatGPT' the common application for LLM requests. However, jailbreak prompts tend to include a reference to the parent company OpenAI, perhaps reflecting an explicit discussion of the corporate policy to be overridden.
In addition, jailbreak prompts can include a specific reference to ethics in their appeal to modify the output of a model. 

\section{Discussion}

We demonstrate that fine-tuning of BERT model to distinguish between jailbreak and non-jailbreak prompts provides the
best overall performance by multiple metrics. We then estimated the performance of this approach on novel jailbreaks
by deliberately withholding all jailbreaks of a certain type from the training process and testing on them alone. This
experiment indicated that performance will likely drop for some types of jailbreak, depending on their similarity with other
jailbreak strategies. Finally, we analysed the key words that appear to distinguish between jailbreak and non-jailbreak
prompts. This work suggested that terms invoking explicit reference to corporate policy or the alignment goals of the model developers feature strongly in jailbreaking strategies.

Many of the techniques used in industry for jailbreak detection and alleviation remain opaque. However, we see that our approach provides a deeper analysis of jailbreak detection with machine learning. In particular, it demonstrates the utility of the BERT model, and the value of evaluating performance on different types of jailbreak. Our semantic analysis of keywords suggests future work can focus on developing jailbreak detection proxy signals such as the degree to which a prompt involves reflexive language related to the model developers and their alignment goals.

\section{Conclusion}

We have observed that optimal detection of jailbreak prompts is achieved by fine tuning a BERT model for classification. This approach outperforms previous approaches to this task using a wider variety of NLP features. Our additional analysis suggests that performance on novel jailbreak prompts can drop, however the amount of change will depend on the nature of the new prompting strategy.

Analysing the keywords used in the jailbreak prompting suggests a higher level of reflexivity in jailbreak prompts. For example, jailbreaks ask the model to explicitly consider the intentions or ethics of the model's parent company when engaging in subversive requests. This suggests that higher order self-referential elements of a prompt be taken on as a theme for further analysis and feature engineering. 

\subsection*{Availability of code and data}

All code and data will be released upon acceptance for publication.

\bibliographystyle{splncs04}
\bibliography{refs}

\begin{thebibliography}{10}
\providecommand{\url}[1]{\texttt{#1}}
\providecommand{\urlprefix}{URL }
\providecommand{\doi}[1]{https://doi.org/#1}

\bibitem{alon2023detectinglanguagemodelattacks}
Alon, G., Kamfonas, M.: Detecting language model attacks with perplexity
  (2023), \url{https://arxiv.org/abs/2308.14132}

\bibitem{Asgari2024.09.12.24313556}
Asgari, E., Monta{\~n}a-Brown, N., Dubois, M., Khalil, S., Balloch, J.,
  Pimenta, D.: A framework to assess clinical safety and hallucination rates of
  llms for medical text summarisation. medRxiv  (2024).
  \doi{10.1101/2024.09.12.24313556},
  \url{https://www.medrxiv.org/content/early/2024/09/13/2024.09.12.24313556}

\bibitem{Austin2021ProgramSW}
Austin, J., Odena, A., Nye, M., Bosma, M., Michalewski, H., Dohan, D., Jiang,
  E., Cai, C.J., Terry, M., Le, Q.V., Sutton, C.: Program synthesis with large
  language models. ArXiv  \textbf{abs/2108.07732} (2021),
  \url{https://api.semanticscholar.org/CorpusID:237142385}

\bibitem{bhargava2024whatsmagicwordcontrol}
Bhargava, A., Witkowski, C., Looi, S.Z., Thomson, M.: What's the magic word? a
  control theory of llm prompting (2024),
  \url{https://arxiv.org/abs/2310.04444}

\bibitem{NEURIPS2020_1457c0d6}
Brown, T., Mann, B., Ryder, N., Subbiah, M., Kaplan, J.D., Dhariwal, P.,
  Neelakantan, A., Shyam, P., Sastry, G., Askell, A., Agarwal, S.,
  Herbert-Voss, A., Krueger, G., Henighan, T., Child, R., Ramesh, A., Ziegler,
  D., Wu, J., Winter, C., Hesse, C., Chen, M., Sigler, E., Litwin, M., Gray,
  S., Chess, B., Clark, J., Berner, C., McCandlish, S., Radford, A., Sutskever,
  I., Amodei, D.: Language models are few-shot learners. In: Larochelle, H.,
  Ranzato, M., Hadsell, R., Balcan, M., Lin, H. (eds.) Advances in Neural
  Information Processing Systems. vol.~33, pp. 1877--1901. Curran Associates,
  Inc. (2020),
  \url{https://proceedings.neurips.cc/paper_files/paper/2020/file/1457c0d6bfcb4967418bfb8ac142f64a-Paper.pdf}

\bibitem{Chandra2022}
Chandra, R., Kulkarni, V.: Semantic and sentiment analysis of selected bhagavad
  gita translations using bert-based language framework. IEEE Access
  \textbf{10},  21291--21315 (2022). \doi{10.1109/ACCESS.2022.3152266}

\bibitem{Chandra2022PLOS}
Chandra, R., Ranjan, M.: Artificial intelligence for topic modelling in hindu
  philosophy: Mapping themes between the upanishads and the bhagavad gita. PLOS
  ONE  \textbf{17}(9),  1--34 (09 2022). \doi{10.1371/journal.pone.0273476},
  \url{https://doi.org/10.1371/journal.pone.0273476}

\bibitem{Chandra2021}
Chandra, R., Saini, R.: Biden vs trump: Modeling us general elections using
  bert language model. IEEE Access  \textbf{9},  128494--128505 (2021).
  \doi{10.1109/ACCESS.2021.3111035}

\bibitem{chang2023survey}
Chang, Y., Wang, X., Wang, J., Wu, Y., Yang, L., Zhu, K., Chen, H., Yi, X.,
  Wang, C., Wang, Y., Ye, W., Zhang, Y., Chang, Y., Yu, P.S., Yang, Q., Xie,
  X.: A survey on evaluation of large language models (2023)

\bibitem{chao2023jailbreaking}
Chao, P., Robey, A., Dobriban, E., Hassani, H., Pappas, G.J., Wong, E.:
  Jailbreaking black box large language models in twenty queries (2023)

\bibitem{chao2024jailbreakingblackboxlarge}
Chao, P., Robey, A., Dobriban, E., Hassani, H., Pappas, G.J., Wong, E.:
  Jailbreaking black box large language models in twenty queries (2024),
  \url{https://arxiv.org/abs/2310.08419}

\bibitem{Chen2021EvaluatingLL}
Chen, M., Tworek, J., Jun, H., Yuan, Q., Pond{\'e}, H., Kaplan, J., Edwards,
  H., Burda, Y., Joseph, N., Brockman, G., Ray, A., Puri, R., Krueger, G.,
  Petrov, M., Khlaaf, H., Sastry, G., Mishkin, P., Chan, B., Gray, S., Ryder,
  N., Pavlov, M., Power, A., Kaiser, L., Bavarian, M., Winter, C., Tillet, P.,
  Such, F.P., Cummings, D.W., Plappert, M., Chantzis, F., Barnes, E.,
  Herbert-Voss, A., Guss, W.H., Nichol, A., Babuschkin, I., Balaji, S., Jain,
  S., Carr, A., Leike, J., Achiam, J., Misra, V., Morikawa, E., Radford, A.,
  Knight, M.M., Brundage, M., Murati, M., Mayer, K., Welinder, P., McGrew, B.,
  Amodei, D., McCandlish, S., Sutskever, I., Zaremba, W.: Evaluating large
  language models trained on code. ArXiv  \textbf{abs/2107.03374} (2021),
  \url{https://api.semanticscholar.org/CorpusID:235755472}

\bibitem{Deng_2024}
Deng, G., Liu, Y., Li, Y., Wang, K., Zhang, Y., Li, Z., Wang, H., Zhang, T.,
  Liu, Y.: Masterkey: Automated jailbreaking of large language model chatbots.
  In: Proceedings 2024 Network and Distributed System Security Symposium. NDSS
  2024, Internet Society (2024). \doi{10.14722/ndss.2024.24188},
  \url{http://dx.doi.org/10.14722/ndss.2024.24188}

\bibitem{deng2024multilingual}
Deng, Y., Zhang, W., Pan, S.J., Bing, L.: Multilingual jailbreak challenges in
  large language models. In: The Twelfth International Conference on Learning
  Representations (2024), \url{https://openreview.net/forum?id=vESNKdEMGp}

\bibitem{bert}
Devlin, J., Chang, M., Lee, K., Toutanova, K.: {BERT:} pre-training of deep
  bidirectional transformers for language understanding. CoRR
  \textbf{abs/1810.04805} (2018), \url{http://arxiv.org/abs/1810.04805}

\bibitem{devlin2019bertpretrainingdeepbidirectional}
Devlin, J., Chang, M.W., Lee, K., Toutanova, K.: Bert: Pre-training of deep
  bidirectional transformers for language understanding (2019),
  \url{https://arxiv.org/abs/1810.04805}

\bibitem{JailbreakLens}
Feng, Y., Chen, Z., Kang, Z., Wang, S., Zhu, M., Zhang, W., Chen, W.:
  Jailbreaklens: Visual analysis of jailbreak attacks against large language
  models. CoRR  \textbf{abs/2404.08793} (2024),
  \url{http://dblp.uni-trier.de/db/journals/corr/corr2404.html#abs-2404-08793}

\bibitem{goyal2023survey}
Goyal, S., Doddapaneni, S., Khapra, M.M., Ravindran, B.: A survey of
  adversarial defenses and robustness in nlp. ACM Comput. Surv.
  \textbf{55}(14s) (jul 2023). \doi{10.1145/3593042},
  \url{https://doi.org/10.1145/3593042}

\bibitem{hackett2025bypassingpromptinjectionjailbreak}
Hackett, W., Birch, L., Trawicki, S., Suri, N., Garraghan, P.: Bypassing prompt
  injection and jailbreak detection in llm guardrails (2025),
  \url{https://arxiv.org/abs/2504.11168}

\bibitem{Hanley1982}
Hanley, J.A., McNeil, B.J.: The meaning and use of the area under a receiver
  operating characteristic (roc) curve. Radiology  \textbf{143}(1),  29--36
  (1982). \doi{10.1148/radiology.143.1.7063747}

\bibitem{hawkins+tivey:2024}
Hawkins, J., Tivey, D.: Literature filtering for systematic reviews with
  transformers. In: 2nd International Conference on Communications, Computing
  and Artificial Intelligence (CCCAI 2024). Jeju, Korea (06 2024).
  \doi{https://doi.org/10.1145/3676581.3676582}

\bibitem{iuChatGPTOpenAIEnd2023}
Iu, K.Y., Wong, V.M.Y.: {ChatGPT} by {OpenAI}: {The} {End} of {Litigation}
  {Lawyers}? (Jan 2023). \doi{10.2139/ssrn.4339839},
  \url{https://papers.ssrn.com/abstract=4339839}

\bibitem{ji2024defendinglargelanguagemodels}
Ji, J., Hou, B., Robey, A., Pappas, G.J., Hassani, H., Zhang, Y., Wong, E.,
  Chang, S.: Defending large language models against jailbreak attacks via
  semantic smoothing (2024), \url{https://arxiv.org/abs/2402.16192}

\bibitem{jiang2024surveylargelanguagemodels}
Jiang, J., Wang, F., Shen, J., Kim, S., Kim, S.: A survey on large language
  models for code generation (2024), \url{https://arxiv.org/abs/2406.00515}

\bibitem{10.5555/3618408.3619130}
Korbak, T., Shi, K., Chen, A., Bhalerao, R., Buckley, C.L., Phang, J., Bowman,
  S.R., Perez, E.: Pretraining language models with human preferences. In:
  Proceedings of the 40th International Conference on Machine Learning.
  ICML'23, JMLR.org (2023)

\bibitem{Lee2024}
Lee, D., Xie, S., Rahman, S., Pat, K., Lee, D., Chen, Q.A.: "prompter says": A
  linguistic approach to understanding and detecting jailbreak attacks against
  large-language models. In: Proceedings of the 1st ACM Workshop on Large AI
  Systems and Models with Privacy and Safety Analysis. pp. 77--87. LAMPS '24,
  Association for Computing Machinery, New York, NY, USA (2024).
  \doi{10.1145/3689217.3690618}, \url{https://doi.org/10.1145/3689217.3690618}

\bibitem{leins-etal-2020-give}
Leins, K., Lau, J.H., Baldwin, T.: Give me convenience and give her death: Who
  should decide what uses of {NLP} are appropriate, and on what basis? In:
  Jurafsky, D., Chai, J., Schluter, N., Tetreault, J. (eds.) Proceedings of the
  58th Annual Meeting of the Association for Computational Linguistics. pp.
  2908--2913. Association for Computational Linguistics, Online (Jul 2020).
  \doi{10.18653/v1/2020.acl-main.261},
  \url{https://aclanthology.org/2020.acl-main.261}

\bibitem{li2024semanticmirrorjailbreakgenetic}
Li, X., Liang, S., Zhang, J., Fang, H., Liu, A., Chang, E.C.: Semantic mirror
  jailbreak: Genetic algorithm based jailbreak prompts against open-source llms
  (2024), \url{https://arxiv.org/abs/2402.14872}

\bibitem{liu2024autodangeneratingstealthyjailbreak}
Liu, X., Xu, N., Chen, M., Xiao, C.: Autodan: Generating stealthy jailbreak
  prompts on aligned large language models (2024),
  \url{https://arxiv.org/abs/2310.04451}

\bibitem{liu2024datasets}
Liu, Y., Cao, J., Liu, C., Ding, K., Jin, L.: Datasets for large language
  models: A comprehensive survey. arXiv preprint arXiv:2402.18041  (2024)

\bibitem{liu2023prompt}
Liu, Y., Deng, G., Li, Y., Wang, K., Wang, Z., Wang, X., Zhang, T., Liu, Y.,
  Wang, H., Zheng, Y., et~al.: Prompt injection attack against llm-integrated
  applications. arXiv preprint arXiv:2306.05499  (2023)

\bibitem{Liu2023}
Liu, Y., Deng, G., Xu, Z., Li, Y., Zheng, Y., Zhang, Y., Zhao, L., Zhang, T.,
  Liu, Y.: Jailbreaking chatgpt via prompt engineering: An empirical study.
  arXiv pp. 1--10 (5 2023)

\bibitem{luo2024jailbreakv28kbenchmarkassessingrobustness}
Luo, W., Ma, S., Liu, X., Guo, X., Xiao, C.: Jailbreakv-28k: A benchmark for
  assessing the robustness of multimodal large language models against
  jailbreak attacks (2024), \url{https://arxiv.org/abs/2404.03027}

\bibitem{JMAI9336}
Mondillo, G., Colosimo, S., Perrotta, A., Frattolillo, V., Indolfi, C., del
  Giudice, M.M., Rossi, F.: Jailbreaking large language models: navigating the
  crossroads of innovation, ethics, and health risks. Journal of Medical
  Artificial Intelligence  \textbf{8}(0) (2024),
  \url{https://jmai.amegroups.org/article/view/9336}

\bibitem{nazir2024langtest}
Nazir, A., Chakravarthy, T.K., Cecchini, D.A., Khajuria, R., Sharma, P., Mirik,
  A.T., Kocaman, V., Talby, D.: Langtest: A comprehensive evaluation library
  for custom llm and nlp models. Software Impacts  \textbf{19},  100619 (2024)

\bibitem{NEURIPS2022_b1efde53}
Ouyang, L., Wu, J., Jiang, X., Almeida, D., Wainwright, C., Mishkin, P., Zhang,
  C., Agarwal, S., Slama, K., Ray, A., Schulman, J., Hilton, J., Kelton, F.,
  Miller, L., Simens, M., Askell, A., Welinder, P., Christiano, P.F., Leike,
  J., Lowe, R.: Training language models to follow instructions with human
  feedback. In: Koyejo, S., Mohamed, S., Agarwal, A., Belgrave, D., Cho, K.,
  Oh, A. (eds.) Advances in Neural Information Processing Systems. vol.~35, pp.
  27730--27744. Curran Associates, Inc. (2022),
  \url{https://proceedings.neurips.cc/paper_files/paper/2022/file/b1efde53be364a73914f58805a001731-Paper-Conference.pdf}

\bibitem{parciak2024schemamatchinglargelanguage}
Parciak, M., Vandevoort, B., Neven, F., Peeters, L.M., Vansummeren, S.: Schema
  matching with large language models: an experimental study (2024),
  \url{https://arxiv.org/abs/2407.11852}

\bibitem{peng2024rapidresponsemitigatingllm}
Peng, A., Michael, J., Sleight, H., Perez, E., Sharma, M.: Rapid response:
  Mitigating llm jailbreaks with a few examples (2024),
  \url{https://arxiv.org/abs/2411.07494}

\bibitem{Preti12024}
Preti1, D., Giannone1, C., Favalli1, A., Romagnoli1, R.: Automatic
  summarization of legal texts, extractive summarization using llms. In:
  Proceedings of the 4th National Conference on Artificial Intelligence
  (Ital-IA 2024). Naples, Italy (May 2024)

\bibitem{qi2023visual}
Qi, X., Huang, K., Panda, A., Henderson, P., Wang, M., Mittal, P.: Visual
  adversarial examples jailbreak aligned large language models. Proceedings of
  the AAAI Conference on Artificial Intelligence  \textbf{38},  21527--21536
  (03 2024). \doi{10.1609/aaai.v38i19.30150}

\bibitem{ran2024jailbreakeval}
Ran, D., Liu, J., Gong, Y., Zheng, J., He, X., Cong, T., Wang, A.:
  Jailbreakeval: An integrated toolkit for evaluating jailbreak attempts
  against large language models (2024)

\bibitem{robey2023smoothllm}
Robey, A., Wong, E., Hassani, H., Pappas, G.J.: Smoothllm: Defending large
  language models against jailbreaking attacks (2023)

\bibitem{sapkota2025aiagentsvsagentic}
Sapkota, R., Roumeliotis, K.I., Karkee, M.: Ai agents vs. agentic ai: A
  conceptual taxonomy, applications and challenges (2025),
  \url{https://arxiv.org/abs/2505.10468}

\bibitem{SCBSZ24}
Shen, X., Chen, Z., Backes, M., Shen, Y., Zhang, Y.: {``Do Anything Now'':
  Characterizing and Evaluating In-The-Wild Jailbreak Prompts on Large Language
  Models}. In: {ACM SIGSAC Conference on Computer and Communications Security
  (CCS)}. ACM (2024)

\bibitem{wang2024decodingtrust}
Wang, B., Chen, W., Pei, H., Xie, C., Kang, M., Zhang, C., Xu, C., Xiong, Z.,
  Dutta, R., Schaeffer, R., Truong, S.T., Arora, S., Mazeika, M., Hendrycks,
  D., Lin, Z., Cheng, Y., Koyejo, S., Song, D., Li, B.: Decodingtrust: A
  comprehensive assessment of trustworthiness in gpt models (2024),
  \url{https://arxiv.org/abs/2306.11698}

\bibitem{Wasti:2024}
Wasti, S.M., Pu, K.Q., Neshati, A.: Large language user interfaces: Voice
  interactive user interfaces powered by llms. In: Arai, K. (ed.) Intelligent
  Systems and Applications. pp. 639--655. Springer Nature Switzerland, Cham
  (2024)

\bibitem{NEURIPS2023_fd661313}
Wei, A., Haghtalab, N., Steinhardt, J.: Jailbroken: How does llm safety
  training fail? In: Oh, A., Naumann, T., Globerson, A., Saenko, K., Hardt, M.,
  Levine, S. (eds.) Advances in Neural Information Processing Systems. vol.~36,
  pp. 80079--80110. Curran Associates, Inc. (2023),
  \url{https://proceedings.neurips.cc/paper_files/paper/2023/file/fd6613131889a4b656206c50a8bd7790-Paper-Conference.pdf}

\bibitem{xu2024comprehensivestudyjailbreakattack}
Xu, Z., Liu, Y., Deng, G., Li, Y., Picek, S.: A comprehensive study of
  jailbreak attack versus defense for large language models (2024),
  \url{https://arxiv.org/abs/2402.13457}

\bibitem{yu2024gptfuzzerredteaminglarge}
Yu, J., Lin, X., Yu, Z., Xing, X.: Gptfuzzer: Red teaming large language models
  with auto-generated jailbreak prompts (2024),
  \url{https://arxiv.org/abs/2309.10253}

\bibitem{zeng2024autodefensemultiagentllmdefense}
Zeng, Y., Wu, Y., Zhang, X., Wang, H., Wu, Q.: Autodefense: Multi-agent llm
  defense against jailbreak attacks (2024),
  \url{https://arxiv.org/abs/2403.04783}

\bibitem{Zhang2025}
Zhang, X., Zhang, C., Li, T., Huang, Y., Jia, X., Hu, M., Zhang, J., Liu, Y.,
  Ma, S., Shen, C.: Jailguard: A universal detection framework for prompt-based
  attacks on llm systems. ACM Trans. Softw. Eng. Methodol.  (Mar 2025).
  \doi{10.1145/3724393}, \url{https://doi.org/10.1145/3724393}, just Accepted

\bibitem{zhuang2023toolqa}
Zhuang, Y., Yu, Y., Wang, K., Sun, H., Zhang, C.: Toolqa: A dataset for llm
  question answering with external tools. Advances in Neural Information
  Processing Systems  \textbf{36},  50117--50143 (2023)

\bibitem{Zou2023UniversalAT}
Zou, A., Wang, Z., Kolter, J.Z., Fredrikson, M.: Universal and transferable
  adversarial attacks on aligned language models. ArXiv
  \textbf{abs/2307.15043} (2023),
  \url{https://api.semanticscholar.org/CorpusID:260202961}

\end{thebibliography}

\end{document}